# Belief–reality separation lives in routing over a shared value slot in language models

**Oliver Steele**
steele@osteele.com
Tsinghua University
NYU Shanghai

**Jiangtao Wen**
jw9263@nyu.edu
NYU Shanghai

**Yuxing Han**
yuxinghan@sz.tsinghua.edu.cn
Tsinghua University

2026-07-08

**Abstract**

Capable language models hold what a character believes apart from what is true: told "Anna believes the cup is blue; in reality it is red," they answer *blue* about Anna and *red* about the world. Where in the computation does that separation live? We show it rests on two separable mechanisms at two positions. A generic *value slot* binds the attributed value. A *router* at the query position selects which frame, the character's belief or reality, a query reads out. Two routes fill the slot: an *asserted* belief, whose value the text supplies, binds in directly; a *derived* belief, whose value must be inferred from what the character could see, arrives by a visibility-gated lookback. A subspace trained on either route steers the other, and only the derived route depends on described visibility. The slot itself carries no belief–reality tag: intervening on it moves a reality readout as strongly as a belief one. The separation lives instead in a dissociated pair of routing subspaces, which flip a query between frames without injecting the donor's value. These results hold across three architectures, on stimuli de-confounded against theory-of-mind-benchmark shortcuts; the behavior itself emerges between 3B and 7B across five model families. This paper develops the single belief–reality axis in depth; a companion paper shows the same slot-and-router format is shared across the other non-actual contexts a sentence can open (counterfactual, fictional, temporal).

## 1 Introduction

Belief is where language most visibly departs from reality. A model trained to continue text about what *is* must also track what merely seems so to someone, and keep the two from contaminating each other: told "Anna believes the cup is blue; in reality it is red," a capable language model answers *blue* when asked what Anna thinks and *red* when asked the fact. This paper asks where that separation lives in the computation, and finds it rests on two mechanisms, distinguished by where each acts. A generic *value slot* holds the attributed value. A *router* at the query position selects which frame, the character's belief or reality, is read out. (Two terms, fixed here at the outset: *slot* and *router* name low-rank continuous subspaces by their functional role, storing a value and selecting a frame, not discrete symbols or registers; *belief-space index*, the companion paper's term (Steele et al., 2026), names the same object as *router*, which we use throughout.) Belief–reality separation lives in the routing, not the value encoding, and both mechanisms hold across three architectures (§4.2, §4.4). The zero-shot display of the attribution itself emerges sharply between 3B and 7B across five model families (§4.1), which fixes 7B as the substrate for the mechanistic analysis. Every stimulus is de-confounded against the shortcuts that inflate theory-of-mind benchmarks (§3), so each behavioral split is evidence about representation. The contrast that carries every experiment is a single one: *blue* for belief against *red* for reality.

Put plainly: the model files the attributed value in one place, a slot, like a labelled card, and selects which frame a question reads it under, belief or reality, in another, a switch at the point of asking. The value and the frame are separate. The belief switch and the reality switch are themselves distinct, not two settings of one knob. The rest of the paper localizes each and shows they come apart.

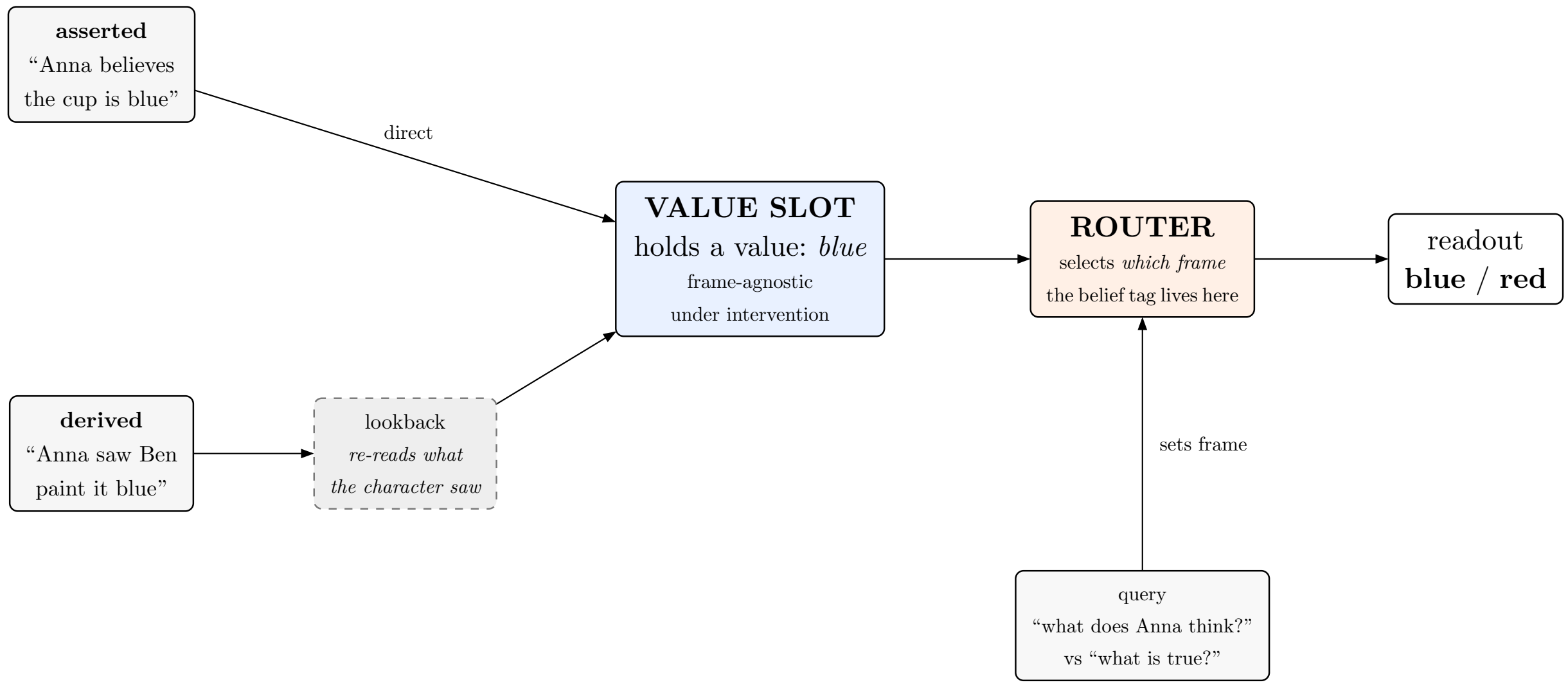

Figure 1: The two mechanisms (each is established in §4.2–4.4). A belief value reaches one generic *value slot* by either of two routes: *asserted* (written directly) or *derived* (a lookback re-reads what the character saw). The slot holds the value but no belief/reality tag; a *router* at the query position selects which frame's value is read out, and is where belief is distinguished from reality.

**The value slot.** The slot is a low-rank subspace of the model's activations that holds the attributed value (here, *blue*). On its own it is not itself the belief tag: intervening on it moves a reality readout exactly as it moves a belief one. It is a generic place a value is written, not a belief-specific store. We conjecture the same slot serves other non-actual contexts, such as counterfactual and depictive reference, and pursue that separately (Steele et al., 2026).

**Two routes into the slot.** The slot is filled either of two ways. An *asserted* belief, whose value the text supplies ("Anna believes the cup is blue"), is written in directly. A *derived* belief withholds its value, which must be inferred from the character's evidence, in all our stimuli what they could see ("Anna saw Ben paint the cup blue but did not see it repainted red"). It reaches the same slot by a derivation prior work characterizes: Prakash et al. (2026) identify a *lookback* that re-reads the character's observed action when the belief is queried. Direct binding and lookback derivation are two routes into one representation (§4.2–4.3 synthesized in §5).

**The router.** The second mechanism is what tags a value as belief rather than reality: a *router* at the query position. A low-rank subspace there selects *which* frame's value a query reads out, dissociated from a matching reality router and decorrelated from the values themselves. (We call the two perspectives a value can be read under, the character's belief and reality, *frames.*) Belief's separation from reality lives in this router, not in the slot it reads. The division of labor with the companion paper (Steele et al., 2026) is fixed by axis: this paper is the depth study of the single belief–reality axis (the two filling routes, visibility gating, and the scale boundary are established here); the companion is the generality study (one router across many space types, and composition, are established there). Each result is established in exactly one of the two papers and imported by the other where needed. A value is re-read from context when queried rather than held in a dedicated store.

Two distinctions organize the evidence. The asserted/derived distinction organizes the first half: if asserted and derived belief are one representation, an intervention that controls one should control the other; and if the derived route is real, derived belief should depend on described visibility in a way asserted belief does not. The belief/reality distinction organizes the second half: if the shared representation carried the frame, an intervention on it would move belief and reality differently. It does not, and the frame selection turns up instead at the routing site.

Our contributions are:

- **One slot, filled two ways.** A low-rank subspace causally controls belief-attributed values, giving full control of the belief readout under interchange. Asserted and derived belief fill that same subspace: a subspace trained on either route steers the other at 0.74–0.97 of its own range on base models (Qwen instruction-tuning sharpens an asymmetry, asserted $\rightarrow$ derived falls to 0.47 while the reverse holds at 0.89; OLMo-2-Instruct is near-symmetric) (§4.2). Derived belief alone depends on described visibility: changing who the believer saw flips it in 99% of cases, while asserted belief is nearly invariant (§4.3). Together these give a two-route account (§5).
- **A frame-agnostic slot, a frame-specific router.** The shared subspace is a generic value slot, not the belief tag: it moves a reality readout as strongly as a belief one (89–95% cross-frame transfer; three architectures, five layers, copy-rule-protected demonstrations). The belief-space index is instead the router at the query position. Belief and reality routers flip a query between frames at their own values, inject none of the donor's color, dissociate under a cross-transfer bar fixed in the analysis plan before the Mistral/OLMo runs, and are near-orthogonal, across Qwen, Mistral, and OLMo-2 (§4.4).
- **A reusable methodology.** A de-confounded stimulus battery (copy-proof two-value designs and order randomization with order-gap discriminators) that survives Ullman-style attacks on ToM benchmarks, together with a completion-template plus few-shot protocol that separates what scale contributes from what instruction-tuning contributes (§3, §5).

The scope of these claims: the scale boundary (§4.1) fixes 7B as the smallest substrate on which the mechanism can be read. Second-order belief, instruction-tuning, and elicitation effects are preliminary, single-substrate in places, and reported in the appendices rather than as primary claims; the most suggestive appendix finding, an order-indexed router family in which nested belief reuses the first-order value slot but mints its own router, is kept there for that reason.

## 2 Related work

### Theory of mind in language models

Whether language models have theory of mind is contested. The standard evaluations are false-belief question-answering benchmarks in the Sally–Anne mold: ToMi (Le et al., 2019), introduced after Le et al. exposed exploitable regularities in earlier bAbI-style ToM benchmarks, extended by SocialIQa-based testing (Sap et al., 2022) and the interactive FANToM (Kim et al., 2023). Kosinski (2024) reports that GPT-family models solve standard false-belief tasks at rates comparable to young children and argues the ability emerged from language modeling; Strachan et al. (2024) likewise compare LLMs and humans across a broad theory-of-mind battery. Concurrent large-sample evidence points the same way as our scale boundary (§4.1): across 41 open-weight models in five families, Trott et al. (2026) find sensitivity to implied knowledge states increasing with scale, without language statistics fully explaining the human effect; their question-versus-completion contrast is the same methodological knob as our completion-template protocol (§3, App. C). Ullman (2023) shows that alterations which leave the answer unchanged for a human collapse that performance, and Shapira et al. (2024) reach a similar conclusion under stress testing, both reading the fragility as shortcut exploitation. Bortoletto et al. (2024) document related brittleness, and Riemer et al. (2025) argue the benchmarks themselves presume a self-consistency these models lack. Two lines respond constructively rather than only critically: Gandhi et al. (2023) generate de-confounded false-belief items from causal templates, and Sclar et al. (2023) raise ToM by maintaining explicit per-character nested belief graphs at decode time, arguing scale alone will not confer a capability they take to be inherently symbolic. Our de-confounded stimuli share the first aim; our finding that nested belief reuses the first-order slot (App. B) bears on the second: the recursion the graph-builder imposes is, in the model, a flat re-application of one-level machinery. This work is almost entirely behavioral. We design every stimulus to defeat the shortcuts the Ullman critique exposes (recency, copying, and canonical task templates), so a behavioral split we report is evidence about representation.

### Mechanisms of belief and entity tracking

Prakash et al. (2026) give the mechanistic account of belief tracking we build on: a lookback that re-reads a character's observed action and binds it into a character-object-state structure, gated by what the character could see. This descends from work on entity binding and state tracking, where a low-rank binding identifier ties an attribute to an entity (Feng and Steinhardt, 2024; Dai et al., 2024) and models maintain entity state across a discourse Kim and Schuster, 2023; Wu et al. (2025a) give the developmental version of the same account, a transformer learning to use the residual stream as an addressable memory with attention heads routing bindings across positions. Retrieval heads (Wu et al., 2024) supply the attentional read that a lookback uses, and the just-in-time task representations of Li et al. (2025), decodable throughout the context but active only at particular tokens, show the same re-read-on-demand profile our re-read-when-queried slot follows. In form, the low-rank subspaces we manipulate resemble the function vectors of Todd et al. (2024) (a compact vector extracted from activations that triggers a role robust to context), but ours carries a discourse frame, not an in-context task. A separate representational line shows a character's belief is linearly decodable and causally steerable (Zhu et al., 2024), while Wu et al. (2025b) identify sparse parameter patterns whose perturbation degrades ToM behavior. Bortoletto et al. (2024) additionally report a layer profile, oracle (reality-aligned) belief representations forming in the earliest layers while protagonist beliefs emerge at intermediate depth, which is consistent with the two-locus picture here, value binding upstream of query-position frame selection, though their probes do not separate the two loci. Prior work thus supplies the derivation (the lookback), causal steering evidence, and the decodability of belief. Zhu et al. decode and manipulate belief states directly; our contribution is the separation of the value slot from the belief/reality router, the cross-frame causal compactness of both loci, and the asserted route that reaches the same slot with no lookback at all. In Prakash et al.'s terms, the router may be close to the source-pointer side of the lookback system, but our current intervention is at the residual-stream query representation, not the QK path; a QK-level pointer test is needed before identifying them. The reality readout we isolate is adjacent to the linear "truth direction" recovered by unsupervised probing (Burns et al., 2023) and by mass-mean probing with causal intervention (Marks and Tegmark, 2024) we differ in treating reality not as a standalone truth axis but as one frame that a query-position router (§4.4) selects among. Our claims are operational rather than adequacy-theoretic: a philosophical literature sets criteria (accuracy, coherence, uniformity, use) for when an internal state counts as a *belief* (Herrmann and Levinstein, 2025), while probe-based criteria that rely on coherence alone can be fragile (Levinstein and Herrmann, 2025). A router that selects the belief frame does not settle those conditions, and we take no position on them; what is localized here is causal structure over readouts, whatever its doxastic status.

# 3 Methods

## 3.1 Models

Our white-box interventions run on open models: Qwen2.5 base at 3B, 7B, 14B, and 32B, Qwen2.5-7B-Instruct (App. A App. B), and Mistral-7B and OLMo-2-7B for architecture generality. The competence boundary of App. A additionally reads larger instruction-tuned and frontier models behaviorally through provider APIs, where the weights needed for intervention are unavailable.

## 3.2 Stimuli

Three minimal-pair templates, each over a small fixed color set and short object and character names, isolate one reported-attitude frame-tracking phenomenon: who is said to believe what, read out as a color completion. They do not test action prediction from belief or broader folk-psychological reasoning. In every case the model answers with a color, so one readout and one intervention serve all three. Color is a uniform readout, not the object of study: the attributed value is arbitrary, and §4.4 replicates the value-slot result with materials in place of colors.

*Selective visibility* (§4.1): two characters each paint an object a color, an observer sees one but not the other, and the observer's belief is queried.

> Anna paints the cup blue. Ben paints the cup red. Carol sees Anna but not Ben. Carol thinks the cup is _____

The seen painter's color is correct (here, blue).

*Asserted versus derived belief* (§4.2–4.3): asserted belief gives the value; derived belief withholds it, so it must be inferred from visibility as above.

> Anna believes the cup is blue. Anna thinks the cup is _____

Asserted and derived items share a query form and a color answer, so a subspace trained on one can be tested on the other.

*Belief versus reality* (§4.4): a single context asserts a belief color and a distinct reality color for one object, clause order randomized, and three query forms read it out.

> Anna believes the cup is blue. In reality the cup is red. {Anna thinks / In fact / The} the cup is _____

The belief query's target is the believed color, the reality query's the true color; the bare query is a diagnostic (it defaults to reality). On base models, whose zero-shot reality readout fails through recency, a few-shot completion prefix restores both readouts; the demonstrations use different verbs and templates than the test items, so a clause-keyed copy rule learned from them cannot transfer (accuracy is unchanged under this decorrelation).

*Second-order false belief* (App. A): one character holds a belief, a second knows the truth, and the model is asked what the second thinks the first believes.

> Ben thinks the cup is blue. Anna knows the cup is actually red. Anna thinks Ben thinks the cup is _____

Ben's false belief is correct (blue); the true color (red) is the diagnostic error.

## 3.3 Controlling for shallow strategies

A behavioral score bears on belief only if the shortcuts that inflate theory-of-mind benchmarks (Ullman, 2023; Shapira et al., 2024) cannot produce it. Two properties of the stimuli remove the two shortcuts that otherwise pass.

- *Recency.* Clause order is randomized independently of the manipulated variable (which character is visible, or which clause holds the true value), so echoing the last-mentioned color is separable from tracking the right one. We report accuracy split by order and read a large order gap as recency, not competence.
- *Copying.* The derived and false-belief items present two distinct colors, so returning the single mentioned value cannot pass; chance between the two present values is 0.50. A copy-baseline confirms the design: a 3B model, which cannot track second-order belief, scores below chance (0.28) on the two-value items through a reality-default pull, yet near-ceiling (0.94) on a one-value variant, where copying suffices.

## 3.4 Interventions

Concretely, one intervention runs like this. The model reads "Anna believes the cup is blue" and, asked what Anna thinks, answers *blue*. We take a second item whose value is *red*, copy the one low-rank slice of internal state our method identifies from the red item into the blue item at the moment of answering, and leave everything else untouched; the answer flips to *red*. That single swap is the experiment in miniature: if copying one slice turns blue into red, the value lives in that slice, and the rest of the paper puts the same question to the belief-versus-reality frame.

The method is Distributed Alignment Search (Geiger et al., 2024; Wu et al., 2023a): it finds the low-rank subspace whose interchange, at the answer token, best moves the readout to the donor's value, the model otherwise frozen. A subspace *controls* a readout if this swap moves the answer to the donor's value. We report *transfer* as the fraction of a subspace's own controllable range it recovers on a second readout, against a random-subspace floor; the floor is zero on the value-token transfer of §4.2. For the router in §4.4, the cross-leg ratio is the residual effect of the wrong-frame router divided by that leg's own-frame routing effect, after subtracting the matched random-subspace floor. All intervention scores reported in the tables are on held-out item tuples disjoint from the DAS optimization items; table captions give the held-out counts (72 for asserted/derived transfer, 90 for router transfer, and 96 for frame-agnostic value-slot transfer). The small template universe is a limitation, so we treat these as controlled mechanistic tests rather than broad behavioral benchmarks.

A subspace that satisfies an interchange objective need not be the model's own mechanism; it can be an expressive fit (Sutter et al., 2025). Three choices guard against that reading. The map is kept a linear orthonormal subspace. Every effect is reported as a margin over the floors above. And the interpretation rests on structure a mere fit would not produce, not on interchange efficacy alone: a subspace trained on belief also moves reality (§4.4), asserted and derived routes fill one slot (§4.2), and the belief and reality routers dissociate (§4.4). A clean-pass projection ablation (§4.4) confirms the routers are used on-distribution, not merely efficacious under interchange.

# 4 Results

## 4.1 Belief attribution from described visibility is architecture-general and emerges with scale

We first fix the substrate. The behavior the mechanism explains is attributing to a character the belief warranted by what that character could see, a documented behavioral bottleneck (Jung et al., 2024). It appears sharply with scale and is not specific to one model family (Figure 2): three architectures fail at 3B, by recency or unreliability, while three solve it cleanly and order-independently at 7B. The same-family crossing is shown for Qwen, and frontier models (GPT-4o, Gemini-2.5) solve it without error (App. C). Action order is randomized independently of visibility, so tracking the seen agent is separable from echoing the last-mentioned color. 7B is therefore the smallest openly-runnable scale at which the mechanism can be studied, and every intervention below runs there or above.

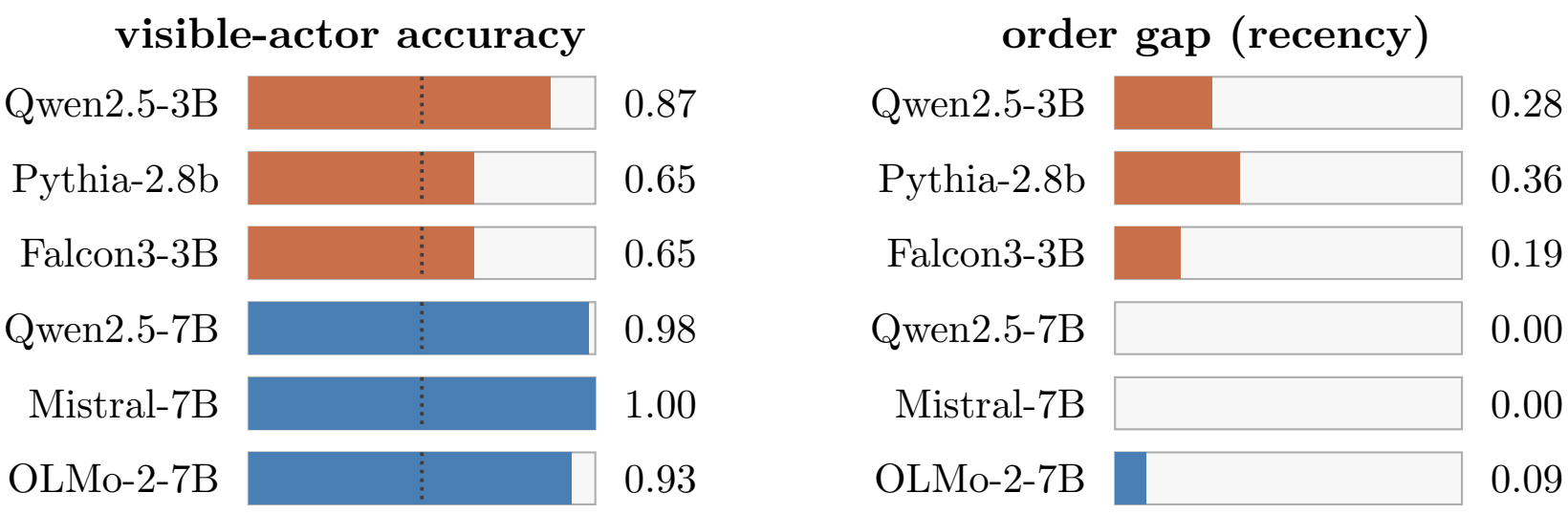

Figure 2: Selective-visibility accuracy (chance 0.50, dotted line) and order gap, the accuracy difference between the seen painter acting last versus first; a large gap indicates recency rather than tracking. Warm bars are 3B models, blue bars 7B. The three 3B models fail order-independence (Qwen2.5-3B reaches 0.87 only with a 0.28 order gap); the three 7B models solve the task order-independently. Chance is 0.25 over the color bank and 0.50 between the two present colors. n = 200 per open model, 60 per API model.

## 4.2 Asserted and derived belief share a binding subspace

We train one DAS subspace to control the asserted-belief readout and another the derived-belief readout, then measure each subspace's control over the other. A random subspace flips neither readout, giving a zero baseline; each trained subspace fully controls its own readout.

The cross-transfers are large (Figure 3). On the base models, a subspace trained only on asserted belief steers derived belief across most of its own range, and the derived subspace steers asserted belief as well or better: to a first approximation, the two forms of belief are one representation. The two learned eight-dimensional subspaces share a roughly three-dimensional core (the top three principal-angle cosines are 0.53 to 0.59, the rest below 0.1), so the large transfer rides on that shared core, with a derivation-specific remainder. Instruction-tuning sharpens the asymmetry: on Qwen2.5-7B-Instruct the derived subspace still steers asserted belief fully, while the asserted subspace steers derived well below its base rate (App. E). This is what one expects if tuning, which makes the model derive belief from visibility more reliably (§4.3), puts more into the derived representation: the shared core persists while the derivation-specific component grows. Asserted and derived belief thus draw on one subspace read from two directions, nearly interchangeable in the base models and more asymmetric once tuning enlarges the derivation step. OLMo-2-7B-Instruct replicates the shared subspace near-symmetrically on a third architecture (Figure 3), without the Qwen instruction-tuned asymmetry (App. E).

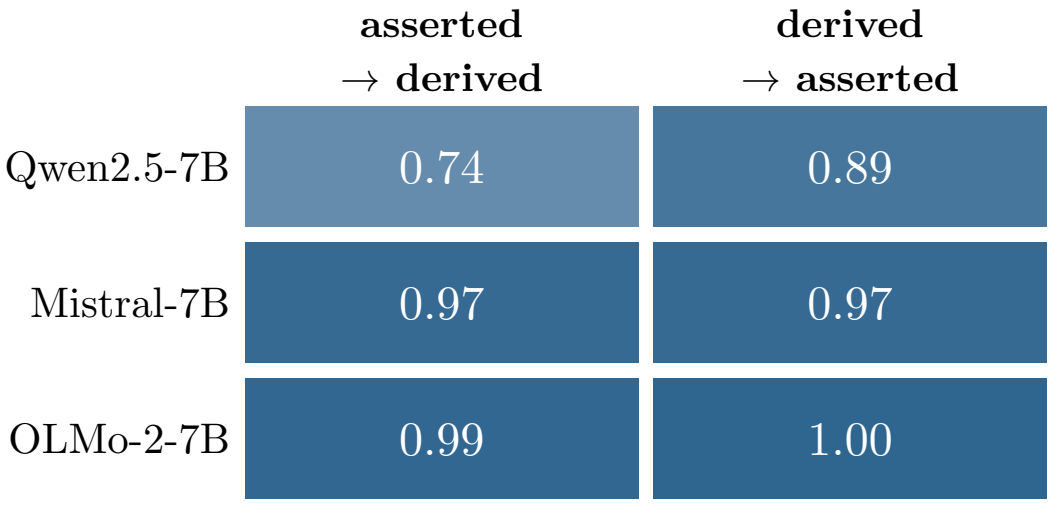

Figure 3: Cross-transfer between the asserted-belief and derived-belief subspaces: the fraction of a subspace's own controllable range that it recovers on the other readout (0 = random-subspace baseline, 1 = full control of its own readout). Base models share the subspace near-symmetrically; Qwen instruction-tuning sharpens the asymmetry, while OLMo-2-Instruct is near-symmetric. Base rows are completion prompts; the OLMo-2 row is instruction-tuned completion. n = 72 test items per model.

### 4.3 Derived belief has a visibility-derivation step that asserted belief lacks

A shared output representation does not settle whether derived belief carries an extra computation. We test for one with an input intervention that needs no activation access: swap which painter the observer sees. If derived belief is computed from visibility, the swap flips the belief to the now-seen painter's color; if asserted belief is pure binding, appending or altering a visibility clause leaves it unchanged.

The contrast is clean and holds on both architectures. Swapping visibility flips the derived belief in 99% of Qwen2.5-7B cases (98.5% at 14B, 99.5% on Mistral-7B), while an asserted belief is nearly invariant to an appended visibility clause (100% unchanged on Qwen at both sizes, 95% on Mistral; n = 200 per model). Derived belief thus depends causally on described visibility, a premise in the text, in a way asserted belief does not. Together with §4.2, this is a two-route account: derived belief is computed by a visibility-gated lookback and written into the same subspace that asserted belief reaches directly.

A conflict control gives the same interpretation behaviorally, with a narrower evidential status. On gpt-4o-mini ($n = 80$), a visible conflicting paint event after an assertion shifts the belief report to the painted color (0.912), while an invisible conflicting event usually preserves the asserted color (0.838). If the assertion is restated after the visible event, the model returns the asserted color (1.000). The invisible-conflict cell rules out a pure last-color recency account; the restated-assertion cell rules out a rule where visible evidence always overrides assertion. A same-family local Qwen replication has not been run, so this result is used as behavioral route-priority evidence rather than as a mechanistic leg.

### 4.4 The value slot is frame-agnostic; a dissociated router selects the frame

The results so far show one subspace filled by two routes. This section asks what that subspace is: does it carry the *belief* tag, as "belief-space index" would suggest, or only the value? The belief-versus-reality stimulus (§3) holds both a believed and a true color in one context, so the two readouts conflict, and every model tested separates them behaviorally (belief readout 0.82–1.00 with reality-leakage ≤ 0.03).

Under interchange, the value-token subspace does not separate belief from reality at all (Table 1). A subspace trained on the belief readout moves the reality readout at 89–95% of its own range, and the reverse, on every substrate, so the subspace both belief routes fill is a *generic value slot*: it encodes the value, not the frame. The result survives the ways it could be an artifact: it holds at five layers spanning 0.25–0.75 of depth on the 7B model, under demonstrations decorrelated from the test templates, and with the value bank changed from colors to materials (transfer 0.94 at 14B). It is not tied to one property type. The frame-agnostic slot replicates on OLMo-2-7B, a third and architecturally distinct family (belief→ reality 0.89, the reverse 0.85, a three-dimensional shared core over a 0.03 floor), so it is not a Qwen–Mistral idiosyncrasy.

This is a causal claim about the subspace that controls the readout, not a claim that the value token contains no frame information. A linear probe trained directly on belief-clause versus reality-clause color-token states in Qwen2.5-7B-Instruct recovers the frame at balanced accuracy 1.000 in layers 1–28, with a grouped split and shuffled-label controls near chance.

Frame information is present at the token. The intervention result says that the low-rank value slot found by DAS is not the frame-selecting component.

| Substrate | belief clean / reality clean | cross-frame transfer | shared core |
|---|---|---|---|
| Qwen2.5-14B (base) | 0.98 / 1.00 | 0.95 | 3 dims |
| Mistral-7B (base) | 0.96 / 0.97 | 0.89 | 3 dims |
| OLMo-2-7B (base) | 0.90 / 0.97 | 0.87 | 3 dims |

Table 1: The value-token subspace is frame-agnostic: cross-frame transfer (belief subspace moving the reality readout and the reverse, fraction of own range over a random-subspace floor) on five substrates, n = 96 items per cell. The Qwen primary row is the seed-robust 14B substrate; Qwen2.5-7B corroborates it on its gate-cleared few-shot completion sample and across five layers (transfer 0.80–0.97). Demonstrations decorrelated from the test templates leave the 7B result unchanged (0.89–0.92).

The behavioral separation must then live elsewhere, and it does: at the query position. We splice a low-rank component of a *frame anchor*, a separate context asserting a belief (or a reality) about a different object in a *different color*, into the final token of a query. The color mismatch is the discriminating control: if the splice injects a value, the output moves to the anchor's color; if it routes, the output moves to the probe's *own* value for that frame. Routing is what happens. A belief-anchor splice flips a reality-defaulting query to the probe's own believed color, and a reality-anchor splice flips a belief-defaulting query to its own true color, each in 100% of cases with zero anchor-color injection, on every substrate (Figure 4).

The belief router and the reality router are separate mechanisms: neither does the other's job. Swapping the two subspaces across legs recovers at most 0.20 of the effect in the primary table below and at most 0.28 across the additional Qwen rows in App. E (under the 0.30 dissociation bar fixed before the non-Qwen replication runs), and the two are near-orthogonal (top principal cosine 0.19–0.28). One secondary preservation control fails on Qwen and is uninformative on Mistral; App. F traces it to slightly off-manifold anchor contexts and scopes it as a limitation of the anchor technique, not a result.

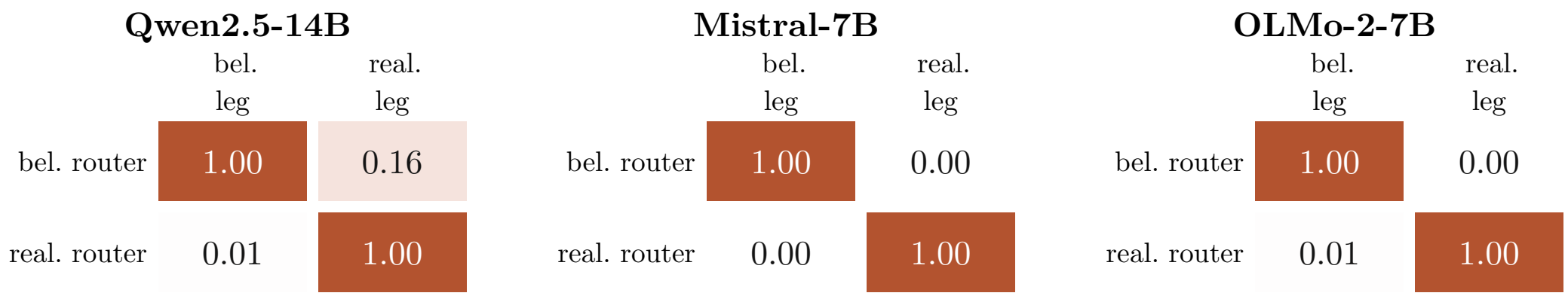

Figure 4: The belief-space index at the routing site, as one router×leg flip-rate matrix per substrate. Rows are the trained router (belief / reality), columns the leg it is applied to; the diagonal is each router flipping its own leg to the probe's own frame value, the off-diagonal is cross-leg leakage. Diagonals are 1.00 and anchor-color injection is 0 everywhere; off-diagonals stay at or below 0.16 against a 0.30 dissociation bar fixed before the non-Qwen replications (n = 90 test items per substrate). Only Qwen shows a cross-leg asymmetry (belief router recovers 0.16 on the reality leg vs the reality router's 0.01 on the belief leg); Mistral and OLMo-2 dissociate symmetrically. Over five seeds Qwen2.5-7B's cross ratios are 0.06 [0.00, 0.11] and 0.08 [0.02, 0.16], with top principal cosine 0.21 to 0.24; the stricter preservation-clean verdict holds on three of five seeds (App. F).

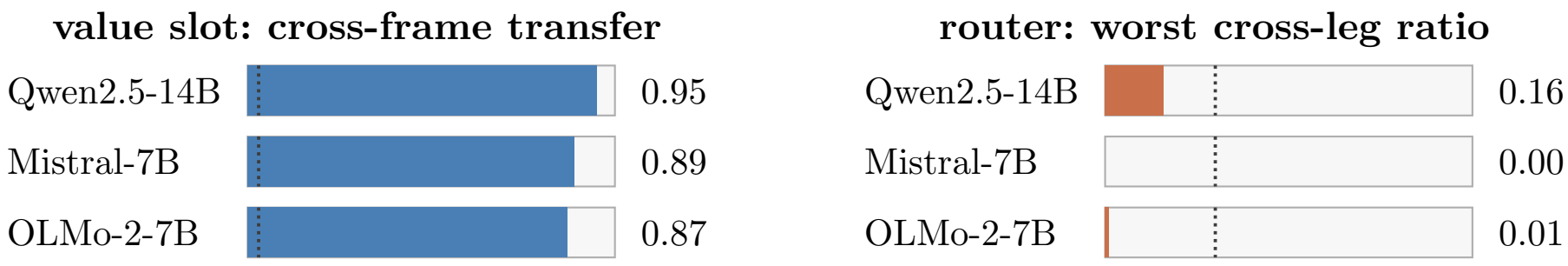

Figure 5: The §4.4 dissection, summarized (values from Table 1 and Figure 4). Left: a value-slot subspace trained on one frame moves the *other* frame's readout at 0.87–0.95 of its own range (dotted line: random-subspace floor, 0.03), so the slot carries no frame tag. Right: a router applied to the wrong leg recovers at most 0.16 of the routing effect (dotted line: the 0.30 dissociation bar fixed in advance; own-leg diagonals are 1.00), so the frame lives in the routing.

The slot/router division is summarized in Figure 5. A probability-margin re-analysis (App. F) confirms the argmax cells are not quantization artifacts of the four-color readout.

A clean-pass projection ablation shows the routers are directions the model uses on its own forward pass, not merely efficacious interchange targets. Projecting a probe's position-matched router out of an un-intervened forward pass collapses its readout toward the competing frame, while the cross-position router and a random rank-8 subspace leave it intact:

| Substrate | belief readout | reality readout | controls |
|---|---|---|---|
| OLMo-2-7B | $0.956 \to 0.000$ | $1.000 \to 0.022$ | cross/random $< 0.01$ |
| Qwen2.5-7B | $0.60 \to 0.00$ | $0.77 \to 0.02$ | cross/random small |

Table 2: Projection ablation of the position-matched router on un-intervened forward passes. Removing the matched router collapses the readout; removing the wrong-position router or a random rank-8 subspace does not. This is the on-pathway evidence against an efficacious-but-off-pathway DAS target.

The belief–reality routing is therefore on-distribution across two substrates, not an efficacious-but-off-pathway illusion (Sutter et al., 2025).

The router also survives a surface-paraphrase control. A router trained on the `believes` surface transfers to `reckons` and `is convinced` on Qwen2.5-7B-Instruct. All three surfaces pass the belief and reality routing gates, and donor-color injection stays at or below 0.014. The router is therefore not just a detector for one lexical query form.

On the Qwen substrates the cross-leg asymmetry is itself informative: the belief router recovers a little of the reality flip while the reality router recovers essentially none, the marked frame's router having some pull toward the default frame but not the reverse. This is clearest at 14B (0.16 versus 0.01) and 7B-Instruct (0.20 versus 0.09); at 7B base a five-seed CI puts the two legs within reach of each other (0.06 / 0.08). On Mistral both cross-leg ratios are 0.00, and OLMo-2′s are near-zero (0.01 / 0.00): symmetric dissociation on both non-Qwen families, no asymmetry. The bare query defaults to reality on every substrate including Mistral and OLMo-2 (0.75–1.00), so the behavioral markedness is general even where the router-level asymmetry is not: the Qwen asymmetry is consistent with reality as the unmarked base, with the reality slot the most isolated subspace in the cross-space geometry, while that router-level signature is Qwen-specific (absent on both Mistral and OLMo-2). Belief–reality separation, then, is implemented as routing over a shared value slot, and "belief-space index" names the router.

# 5 Discussion

Belief in a language model is distinguished by a value slot and a router, filled and selected by separable mechanisms. The two routes into the slot (§4.2–4.3) cast the lookback of Prakash et al. (2026) as the derivation that fills it. They also extend that account in two directions it does not itself reach: the representation the lookback writes into, and the asserted case that needs no lookback at all.

The §4.4 dissection reframes what is special about belief. It is not the slot. The slot carries no belief-versus-reality tag within belief itself. Cross-type generality, the same slot and a shared router across counterfactual, depictive, temporal, and other spaces, is established in the companion paper, not here; we import it for context (Steele et al., 2026). In that companion setting the space-index signal also survives surface-frame variation across nine builder types, so the comparison is not pinned to one belief template; nothing in the present paper's own evidence extends beyond the belief–reality axis. What carries the belief–reality distinction is the *router*: the query-position subspace that selects the belief frame's value over reality's, dissociated from its reality counterpart and decorrelated from the values it routes. A study that intervenes only at the value token finds no belief–reality separation at all, because it is looking at the shared slot. The two loci answer different questions: the slot, what is attributed; the router, whose frame is read.

The reality-default thread runs through the behavioral and mechanistic results. In the shallow-copy control, a 3B model falls below chance on two-value false-belief items by pulling toward reality; in App. A, base 7B's plain second-order completion likewise defaults to the true color; in §4.4, the bare query defaults to reality on every substrate. Qwen's small cross-leg asymmetry is the mechanistic version of the same markedness pattern: the belief router can partially recover the reality flip, while the reality router mostly does not recover the belief flip. The default itself is general, but the router-level asymmetry is Qwen-specific, so we read markedness as a behavioral organizing principle rather than a universal circuit signature.

The recursive appendix extends the same slot/router distinction. Second-order belief reuses the first-order value slot, but the nested frame is selected by its own router on the one substrate where the three-value stimulus is legible. That order-indexed router family is the paper's most suggestive recursive mechanism; it is kept in the appendix because it is single-substrate, not because it is peripheral to the account.

The scale boundary and the elicitation analysis are developed in App. C: zero-shot accuracy changes sharply with instruction-tuning, but few-shot and scaffolded variants show that the boundary mixes capability with elicitation. The same-size jump that locates the 7B substrate therefore does not by itself settle why.

# 6 Conclusion

Belief in a transformer language model is not a single representation but a division of labor between two low-rank subspaces at two positions. A generic value slot holds the attributed value, filled directly for an asserted belief and by a visibility-gated lookback for a derived one. A query-position router, dissociated from its reality counterpart, then selects which frame that value is read under. Belief–reality separation is therefore implemented as routing over a shared value slot, and "belief-space index" names the router: a result that holds across three architectures, and that a value-token-only analysis would miss entirely.

# 7 Limitations and future work

**Coverage.** The mechanistic results are on models of 7B and 14B parameters. The asserted/derived shared subspace (§4.2), the frame-agnostic value slot (§4.4), and the router dissociation (§4.4) are each established on three architectures (Qwen, Mistral-7B, OLMo-2), with the OLMo-2 substrate instruction-tuned at §4.2 and base-with-few-shot at §4.4 coverage by result is mapped in App. D. The behavioral boundary spans more families and sizes but is read through provider APIs for the larger and instruction-tuned models, so quantization and serving differ across points on that curve.

**The second-order boundary is framing-sensitive near 8B.** It is tested on two false-belief framings, unexpected-contents and change-of-location (App. A). It replicates for the models that pass both, so it should be read as robust for both-framing passers rather than as a sharp universal threshold.

**The derivation step is not circuit-localized.** The visibility-derivation step is demonstrated by input intervention, which shows the dependence without localizing the circuit; an activation-level localization, and its relation to the retrieval-head literature (Wu et al., 2024), is left to future work.

**The recursive result is single-model.** App. B's slot-reuse result is on one model in the chat framing, the only framing in which second-order belief is legible at this scale; whether the reuse holds across architectures is open. App. B does find a distinct nested router on one 14B base substrate, so the open router question is cross-architecture replication rather than existence.

**The anchor-splice technique carries an unresolved footnote.** Its secondary preservation control fails on the Qwen runs and is below detection on Mistral (preservation 0.667 versus a 0.633 random baseline), so the off-manifold issue is Qwen-specific rather than resolved (App. F). The per-run intervention verdicts in §4.4 are reported against fixed analysis bars, with this failure disclosed rather than absorbed.

**The §4.4 readout gates are item-sample sensitive at 7B.** Base and instruct 7B clear both clean-accuracy gates on one of five stimulus samples (the cross-frame transfer itself is stable, 0.81–0.92, across all samples). At 14B all five samples clear both gates with cross-frame transfer 0.89–0.96, so the primary substrate carries a replicated range and the 7B rows are single-sample corroboration.

**Two controls are single-family scope guards.** Value-token frame decodability and surface-paraphrase router transfer are each measured on Qwen2.5-7B-Instruct only. They sharpen the claim and reduce template risk, but they are not additional cross-architecture mechanism legs.

# 8 Data and code availability

A controlled stimulus suite for the mental-space constructions studied here is released as the Mental Spaces Corpus (Steele, 2026) (doi:10.57967/hf/9657 https://huggingface.co/datasets/osteele/mental-spaces). Build and validation code is at https://github.com/osteele/mental-spaces.

# References

Matteo Bortoletto, Constantin Ruhdorfer, Lei Shi, and Andreas Bulling. 2024. Brittle Minds, Fixable Activations: Understanding Belief Representations in Language Models. *Findings of EMNLP 2025*.

Collin Burns, Haotian Ye, Dan Klein, and Jacob Steinhardt. 2023. Discovering Latent Knowledge in Language Models Without Supervision. *International Conference on Learning Representations (ICLR)*.

Qin Dai, Benjamin Heinzerling, and Kentaro Inui. 2024. Representational Analysis of Binding in Language Models. *Proceedings of the 2024 Conference on Empirical Methods in Natural Language Processing*:17468–17493.

Jiahai Feng and Jacob Steinhardt. 2024. How Do Language Models Bind Entities in Context?. *International Conference on Learning Representations*.

Kanishk Gandhi, Jan-Philipp Fränken, Tobias Gerstenberg, and Noah D. Goodman. 2023. Understanding Social Reasoning in Language Models with Language Models. *Advances in Neural Information Processing Systems (NeurIPS), Datasets and Benchmarks Track*.

Atticus Geiger, Zhengxuan Wu, Christopher Potts, Thomas Icard, and Noah D. Goodman. 2024. Finding Alignments Between Interpretable Causal Variables and Distributed Neural Representations. *Proceedings of Machine Learning Research*, 236:160–187. CLeaR 2024.

Daniel A. Herrmann and Benjamin A. Levinstein. 2025. Standards for Belief Representations in LLMs. *Minds and Machines*, 35(1):5.

Chani Jung, Dongkwan Kim, Jiho Jin, Jiseon Kim, Yeon Seonwoo, Yejin Choi, Alice Oh, and Hyunwoo Kim. 2024. Perceptions to Beliefs: Exploring Precursory Inferences for Theory of Mind in Large Language Models. *Empirical Methods in Natural Language Processing*.

Hyunwoo Kim, Melanie Sclar, Xuhui Zhou, Ronan Le Bras, Gunhee Kim, Yejin Choi, and Maarten Sap. 2023. FANToM: A Benchmark for Stress-testing Machine Theory of Mind in Interactions. *Proceedings of the 2023 Conference on Empirical Methods in Natural Language Processing (EMNLP)*:14397–14413.

Najoung Kim and Sebastian Schuster. 2023. Entity Tracking in Language Models. *Proceedings of the 61st Annual Meeting of the Association for Computational Linguistics*:3835–3855.

Michal Kosinski. 2024. Evaluating large language models in theory of mind tasks. *Proceedings of the National Academy of Sciences*, 121(45):e2405460121.

Matthew Le, Y-Lan Boureau, and Maximilian Nickel. 2019. Revisiting the Evaluation of Theory of Mind through Question Answering. *Proceedings of the 2019 Conference on Empirical Methods in Natural Language Processing (EMNLP-IJCNLP)*:5872–5877.

Benjamin A. Levinstein and Daniel A. Herrmann. 2025. Still No Lie Detector for Language Models: Probing Empirical and Conceptual Roadblocks. *Philosophical Studies*, 182:1539–1565.

Yuxuan Li, Declan Campbell, Stephanie C. Y. Chan, and Andrew Kyle Lampinen. 2025. Just-in-time and Distributed Task Representations in Language Models. *arXiv preprint arXiv:2509.04466*. Spotlight, NeurIPS 2025 Mechanistic Interpretability Workshop.

Samuel Marks and Max Tegmark. 2024. The Geometry of Truth: Emergent Linear Structure in Large Language Model Representations of True/False Datasets. *Conference on Language Modeling (COLM).*

Nikhil Prakash, Tamar Rott Shaham, Tal Haklay, Yonatan Belinkov, and David Bau. 2024. Fine-Tuning Enhances Existing Mechanisms: A Case Study on Entity Tracking. *International Conference on Learning Representations.*

Nikhil Prakash, Natalie Shapira, Arnab Sen Sharma, Christoph Riedl, Yonatan Belinkov, Tamar Rott Shaham, David Bau, and Atticus Geiger. 2026. Language Models Use Lookbacks to Track Beliefs. *International Conference on Learning Representations.* arXiv:2505.14685v3.

Matthew Riemer, Zahra Ashktorab, Djallel Bouneffouf, Payel Das, Miao Liu, Justin D. Weisz, and Murray Campbell. 2025. Position: Theory of Mind Benchmarks are Broken for Large Language Models. *International Conference on Machine Learning.*

Maarten Sap, Ronan Le Bras, Daniel Fried, and Yejin Choi. 2022. Neural Theory-of-Mind? On the Limits of Social Intelligence in Large LMs. *Proceedings of the 2022 Conference on Empirical Methods in Natural Language Processing (EMNLP)*:3762–3780.

Melanie Sclar, Sachin Kumar, Peter West, Alane Suhr, Yejin Choi, and Yulia Tsvetkov. 2023. Minding Language Models' (Lack of) Theory of Mind: A Plug-and-Play Multi-Character Belief Tracker. *Proceedings of the 61st Annual Meeting of the Association for Computational Linguistics (ACL).*

Natalie Shapira, Mosh Levy, Seyed Hossein Alavi, Xuhui Zhou, Yejin Choi, Yoav Goldberg, Maarten Sap, and Vered Shwartz. 2024. Clever Hans or Neural Theory of Mind? Stress Testing Social Reasoning in Large Language Models. In *Proceedings of the 18th Conference of the European Chapter of the Association for Computational Linguistics (EACL).*

Oliver Steele. 2026. Mental Spaces Corpus. Hugging Face dataset, version 0.1, doi:10.57967/hf/9657.

Oliver Steele, Jiangtao Wen, and Yuxing Han. 2026. One mechanism for many mental spaces: a shared router over a value slot in language models. *arXiv preprint arXiv:2607.10248.*

James W. A. Strachan, Dalila Albergo, Giulia Borghini, Oriana Pansardi, Eugenio Scaliti, Saurabh Gupta, Krati Saxena, Alessandro Rufo, Stefano Panzeri, Guido Manzi, Michael S. A. Graziano, and Cristina Becchio. 2024. Testing Theory of Mind in Large Language Models and Humans. *Nature Human Behaviour*, 8:1285–1295.

Winnie Street, John Oliver Siy, Geoff Keeling, Adrien Baranes, Benjamin Barnett, Michael McKibben, Tatenda Kanyere, and others. 2025. LLMs Achieve Adult Human Performance on Higher-Order Theory of Mind Tasks. *Frontiers in Human Neuroscience.*

Denis Sutter, Julian Minder, Thomas Hofmann, and Tiago Pimentel. 2025. The Non-Linear Representation Dilemma: Is Causal Abstraction Enough for Mechanistic Interpretability?. *Advances in Neural Information Processing Systems.*

Eric Todd, Millicent L. Li, Arnab Sen Sharma, Aaron Mueller, Byron C. Wallace, and David Bau. 2024. Function Vectors in Large Language Models. *International Conference on Learning Representations (ICLR).*

Sean Trott, Samuel Taylor, Cameron Jones, James A. Michaelov, and Pamela D. Rivière. 2026. Language Statistics and False Belief Reasoning: Evidence from 41 Open-Weight LMs. *arXiv preprint arXiv:2602.16085.*

Tomer Ullman. 2023. Large Language Models Fail on Trivial Alterations to Theory-of-Mind Tasks. *arXiv preprint arXiv:2302.08399.*

Wenhao Wu, Yizhong Wang, Guangxuan Xiao, Hao Peng, and Yao Fu. 2024. Retrieval Head Mechanistically Explains Long-Context Factuality. *arXiv preprint arXiv:2404.15574.*

Yiwei Wu, Atticus Geiger, and Raphaël Millière. 2025a. How Do Transformers Learn Variable Binding in Symbolic Programs?. *International Conference on Machine Learning.*

Yufan Wu, Yinghui He, Yilin Jia, Rada Mihalcea, Yulong Chen, and Naihao Deng. 2023b. Hi-ToM: A Benchmark for Evaluating Higher-Order Theory of Mind Reasoning in Large Language Models. *Findings of the Association for Computational Linguistics: EMNLP 2023.*

Yuheng Wu, Wentao Guo, Zirui Liu, Heng Ji, Zhaozhuo Xu, and Denghui Zhang. 2025b. Sensitivity Meets Sparsity: The Impact of Extremely Sparse Parameter Patterns on Theory-of-Mind of Large Language Models. *arXiv preprint arXiv:2504.04238.*

Zhengxuan Wu, Atticus Geiger, Thomas Icard, Christopher Potts, and Noah D. Goodman. 2023a. Interpretability at Scale: Identifying Causal Mechanisms in Alpaca. *Advances in Neural Information Processing Systems*, 36.

Wentao Zhu, Zhining Zhang, and Yizhou Wang. 2024. Language Models Represent Beliefs of Self and Others. *International Conference on Machine Learning.*

# APPENDIX A Second-order false belief requires instruction-tuning and scale

*Appendices A and B report preliminary, single-substrate results demoted from the main text. They are not central to the mechanism established there; they record where the recursive case stands, held to a lower evidential bar.*

The recursive case, where a model must represent one character's belief about another's false belief, is where behavioral ToM studies concentrate (Wu et al., 2023b; Street et al., 2025) and where those critiques are sharpest. We measure it with a copy-proof false-belief stimulus: a character holds a belief, a second character knows the truth, and the model is asked what the second thinks the first believes, with both colors present and clause order randomized so that neither copying nor recency can pass. Chance between the two present values is 0.50.

Zero-shot competence appears only at the conjunction of instruction-tuning and scale (Figure 6). Base models fail the plain completion display at every size tested, up to 32B, reporting the true color rather than the attributed belief; instruction-tuned models below about 7B fail as well. From about 7B, instruction-tuned models pass, and larger and frontier models are error-free. The decisive comparison is same-size: Qwen2.5-7B moves from 0.09 (base) to 0.94 (instruct), which rules out a residual stimulus artifact, since a copying or format confound would not respond to instruction-tuning. This is a display boundary, not a proof that base models lack the underlying computation: App. C shows base 7B/14B can reach ceiling under an explicit scaffold. The smallest openly-runnable substrate for an unscaffolded mechanistic study of the recursive case is therefore an instruction-tuned model of about 7B.

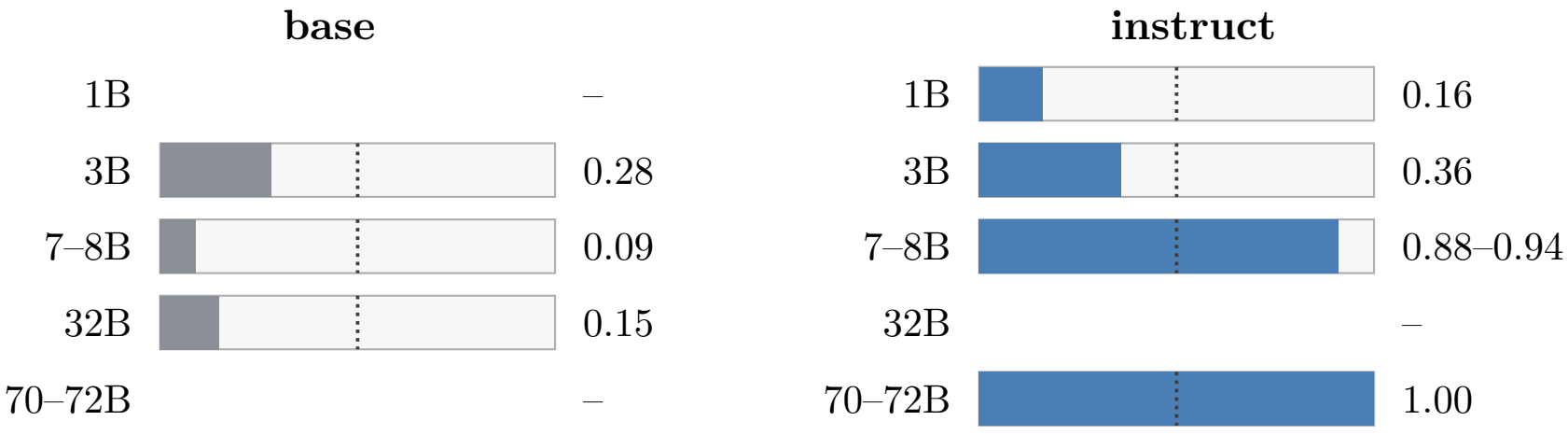

Figure 6: Second-order false-belief accuracy on the copy-proof stimulus (chance 0.50, dotted line; – marks a size not tested in that column). Base models fail at every size; instruction-tuned models cross between 3B and 7B; frontier models score 1.00. Same-size, Qwen2.5-7B rises from 0.09 (base) to 0.94 (instruct). n = 200 per open model, 60 per API model.

A single false-belief framing cannot establish robustness, since the critiques above concern precisely the case where competence on one task template does not survive a change of framing. We re-ran the boundary on a second framing, change-of-location: a character puts an object in one place and leaves, a second character moves it, and the model is asked what the second thinks the first believes. The mention order of the two locations is randomized, so that answering by position is separable from tracking, measured as an order gap (§3, here over the two locations). For the models that clear the boundary the result replicates: Qwen2.5-7B-Instruct passes at 0.97 with a near-zero order gap (0.07), 70B, 72B, and GPT-4o pass without error, and instruction-tuned models at 1B and 3B stay near chance. The second framing also reveals a framing-specific failure the first does not: Llama-3.1-8B-Instruct, which passes the contents framing at 0.88, falls to position-based guessing on change-of-location (0.57 to 0.65 across two seeds, order gap 0.6 to 0.7). The boundary is therefore robust for models that pass both framings and framing-sensitive at the margin near 8B, the size range where a template-specific shortcut is most expected to appear. The order gap, not raw accuracy, is what separates tracking from position-guessing there.

# APPENDIX B Nested belief fills the same slot

We run the interchange on the smallest substrate where the recursive case runs (App. A): does second-order belief reuse the first-order binding subspace or recruit a distinct nested one? We apply the transfer method of §4.2 on Qwen2.5-7B-Instruct. Second-order belief is legible only under the chat framing through which the larger models were measured (App. A): in plain completion the second-order readout sits near chance, but in the chat framing both the first-order and second-order readouts are at ceiling, so each can serve as a clean interchange target. In that framing the transfer is clean and near-symmetric. A subspace trained to control first-order belief steers the second-order readout at 0.90 of its own range over a zero baseline, and the second-order subspace steers first-order belief at 1.00, on a three-dimensional shared core. Second-order belief therefore reuses the first-order binding subspace rather than recruiting a distinct nested one: attributing a belief to another agent writes into the same slot, applied recursively. The router is another matter: on the one substrate where a three-value deception stimulus (one that separates the nested frame's value from the first-order belief) is legible, base Qwen2.5-14B under few-shot completion, the §4.4 method finds the nested frame selected by its *own* routing subspace, dissociated from the first-order router (cross-transfer ratios 0.00 both directions, top principal cosine 0.34, zero anchor-color injection with the continuous check). Second-order belief reuses the slot but not the router; the router family appears to be order-indexed. Parallel compound-space results in the companion paper show the same signature for nested depiction (painting-of-photograph and photograph-of-painting) and for a mixed painting-in-belief frame, so the pattern is not specific to belief. The belief-specific recursive slot result here remains one-model; the companion paper now reports cross-architecture support for the broader nested and mixed-compound signature.

# APPENDIX C Scale, elicitation, and instruction-tuning

The capability appears sharply with scale and is not specific to one model family (Figure 2). At 3B, three architectures fail in three different ways: Qwen2.5-3B leans on recency, Pythia-2.8b is pure recency (its accuracy halving when the seen painter acts first), and Falcon3-3B is unreliable rather than recency-bound. At 7B, three architectures solve it cleanly and order-independently, and frontier models (GPT-4o, Gemini-2.5) solve the task without error. Three families fail at 3B and three solve it at 7B, with the same-family crossing shown for Qwen, so the boundary tracks scale rather than one family's quirk; 7B is the smallest openly-runnable scale at which the mechanism can be studied.

The competence boundary, read together with the elicitation results, gives a sharper picture of what instruction-tuning contributes than "tuning installs the capability." The two factors dissociate, and they dissociate differently at the two orders.

At first order, what scale gates at 3B is the belief-versus-reality conflict, not attribution itself. Visibility-based attribution appears between 3B and 7B across model families (§4.1), but the same few-shot protocol that leaves that boundary intact for the conflict task largely dissolves it for the visibility task (3B reaches 0.87, order-independent, under four demonstrations), so §4.1's zero-shot boundary is partly elicitation. The conflict is different: the demonstrations that lift base 7B to 0.95 instead teach 3B an order-dependent strategy, its belief readout degrading under them.

The 3B cell is thus the cleanest capability-versus-elicitation dissociation in our data. Within one model and one protocol, visibility-based attribution is present but unelicited, while conflict-holding is absent; the same demonstrations elicit the one and counterfeit the other.

At first order, tuning changes what is elicited, not what is represented. Base models fail the belief-versus-reality readouts through recency, but four demonstrations, in wordings a copy rule cannot exploit, restore them to 0.95–1.00. And the tuned model's value-slot subspace matches the base model's at the seed-stability ceiling of the measurement, so the machinery the tuned model uses is the base model's, consistent with tuning enhancing existing mechanisms rather than installing new ones (Prakash et al., 2024).

At second order the zero-shot conjunction stands behaviorally (App. A), but it is an elicitation gap, not absent capacity. Plain few-shot completion does not lift base second-order accuracy (0.11 at 7B; an order-driven 0.37 at 14B), but a reasoning-walk scaffold that names the recursion lifts it from floor to ceiling on both base sizes (1.00, order-robust), with first-order at ceiling under the identical scaffold. Representations a scaffold can drive to ceiling are present, not absent: the zero-shot failure is a failure to cue the recursive composition, not to hold it, consistent with §4.4, where base models already route a directly-asserted nested attribution when cued (0.96). A minimal scaffold supplying only the two steps (B's belief, then A's view of it, with no answer-selection rule stated) lifts both base sizes to ceiling as well (0.99/1.00), so base composes the recursion when cued rather than executing a stated rule; whether it composes with no scaffold at all is untested. A base model that fails the behavior is in either case the wrong substrate for mechanistic study of it; the same-size jump (0.09 to 0.94 at 7B) locates the substrate without yet settling why.

## APPENDIX D Coverage by result

| Result | Qwen 3–14B | Mistral-7B | Other open families | Frontier (API) |
|---|---|---|---|---|
| §4.1 visibility→ belief emergence *(behav.)* | ✓ | ✓ | ✓ Pythia, Falcon3, OLMo | ✓ |
| §4.2 asserted/derived share a subspace *(causal)* | ✓ | ✓ | ✓ OLMo-2 | – |
| §4.3 visibility-derivation step *(causal)* | ✓ | ✓ | – | – |
| App. A second-order boundary *(behav.)* | ✓ | – | ✓ Llama, 1–72B | ✓ |
| App. B nested reuses first-order slot *(causal)* | ✓ *Instruct-7B only* | – | – | – |
| §4.4 frame-agnostic slot *(causal)* | ✓ | ✓ | ✓ OLMo-2 | – |
| §4.4 belief/reality router *(causal)* | ✓ | ✓ | ✓ OLMo-2 | – |

Table 3: Model coverage by result. The causal mechanism (§4.2–4.4) is established across two to three architectures (Qwen, Mistral-7B, OLMo-2), as the rows show; the behavioral boundaries (§4.1, App. A) span five open families plus frontier models; App. B is a single-model result (see Limitations). ✓ tested; – not tested.

## APPENDIX E Additional model rows

The primary tables report one substrate per architecture. The Qwen scale and instruction-tuned rows the main text summarizes are collected here.

| Model | asserted → derived | derived → asserted |
|---|---|---|
| Qwen2.5-14B | 0.74 | 0.79 |
| Qwen2.5-7B-Instruct | 0.47 | 0.89 |

Table 4: Asserted/derived cross-transfer (§4.2), additional Qwen rows. Instruction-tuning sharpens the asymmetry (asserted → derived falls to 0.47).

| Substrate | belief clean / reality clean | cross-frame transfer | shared core |
|---|---|---|---|
| Qwen2.5-7B (base) | 0.82 / 0.98 | 0.92 | 2–3 dims |
| Qwen2.5-7B-Instruct | 0.93 / 0.99 | 0.92 | 3 dims |

Table 5: Frame-agnostic slot (§4.4), additional Qwen rows.

| Substrate | routing (own value) | injection (anchor color) | cross-leg ratio |
|---|---|---|---|
| Qwen2.5-7B-Instruct | 1.00 / 1.00 | 0.00 | 0.09 / 0.20 |
| Qwen2.5-7B (base) | 1.00 / 1.00 | 0.00 | 0.06 / 0.08 |

Table 6: Belief/reality router (§4.4), additional Qwen rows. Diagonals 1.00, injection 0; the cross-leg asymmetry is Qwen-specific (see §4.4).

# APPENDIX F Robustness checks for the router analysis

Two checks referenced from §4.4 are collected here.

**The secondary preservation control.** One secondary control fails, on every substrate: a belief-anchor splice applied redundantly to an already-belief-defaulting query partially disrupts it. A donor-source ladder traces this to the anchor contexts being slightly off-manifold for already-marked queries in Qwen; on Mistral the preservation value is 0.667 against a 0.633 random baseline, so that control is uninformative rather than passed. We keep the preservation issue as a methodological footnote of the anchor technique, not a result; it is disclosed in the per-run verdicts rather than absorbed.

**Probability-margin re-analysis.** A probability-margin re-analysis confirms the §4.4 argmax cells are not quantization artifacts of the four-color readout: the routing diagonals carry softmax probability $\approx 0.99$, the zero-injection cells stay below 0.08 continuously, and the value-token cross-frame transfer is a genuine broad unimodal spread (per-item donor-color probability median $\approx$ mean, no masked subpopulation) rather than a mean concealing structure.